\documentclass[runningheads]{llncs}
\usepackage{xcolor}
\usepackage{graphicx}
\usepackage{hyperref}
\usepackage{floatrow}
\usepackage{amsmath}
\usepackage{wrapfig}
\usepackage{subfigure}

\usepackage{blindtext}

\DeclareRobustCommand*{\ora}{\overrightarrow}

\begin{document}

\title{Comparing Shape-Constrained Regression Algorithms for Data Validation}
\titlerunning{Comparing Shape-Constrained Regression Algorithms for Data Validation}

\authorrunning{Bachinger and Kronberger}

\author{
  Florian Bachinger\inst{1,2}\orcidID{0000-0002-5146-1750}
  \and
  Gabriel Kronberger\inst{1}\orcidID{0000-0002-3012-3189}
}

\institute{
  Josef Ressel Center for Symbolic Regression\\
  Heuristic and Evolutionary Algorithms Laboratory\\
  University of Applied Sciences Upper Austria, Hagenberg, Austria\\
  \and
  Institute for Application-oriented Knowledge Processing (FAW)\\
  Johannes Kepler University, Linz, Austria\\
  \email{florian.bachinger@fh-hagenberg.at}
}
\maketitle 
\renewcommand{\thefootnote}{}
\footnotetext{\hspace{-0em}
	 Submitted manuscript to be published in \textit{Computer Aided Systems Theory - EUROCAST 2022: 18th International Conference, Las Palmas de Gran Canaria, Feb. 2022}.
}
\renewcommand\thefootnote{\arabic{footnote}}
\begin{abstract}
  Industrial and scientific applications handle large volumes of data that render manual
  validation by humans infeasible. Therefore, we require automated data validation
  approaches that are able to consider the prior knowledge of domain experts to produce
  dependable, trustworthy assessments of data quality. Prior knowledge is often available
  as rules that describe interactions of inputs with regard to the target e.g. the target
  must be monotonically decreasing and convex over increasing input values. Domain experts
  are able to validate multiple such interactions at a glance. However, existing
  rule-based data validation approaches are unable to consider these constraints. In this
  work, we compare different shape-constrained regression algorithms for the purpose of
  data validation based on their classification accuracy and runtime performance.
\end{abstract}

\section{Introduction}
Modern applications record a staggering amount of data through the application of sensor
platforms. These masses of data render manual validation infeasible and require automated
data validation approaches. Existing rule-based approaches~\cite{Ehrlinger2022} can
detect issues like missing values, outliers, or changes in the distribution of individual
observables. However, they are unable to assess the data quality based on interactions of
multiple observables with regard to a target. For example, they might falsely classify an
outlier as invalid, even though it can be explained by changes in another variable.
Alternatively, an observable might exhibit valid value ranges and distributions, whilst
the error is only detectable in the unexpected interaction with other observables, e.g.
one dependent variable remains of constant value while another changes.

For this purpose, we propose the use of shape constraints (SC) for data validation. We
detail the general idea of SC-based data validation and provide a comparison of three
algorithms: (1) shape-constrained polynomial regression (SCPR)~\cite{HallPhd}, (2)
shape-constrained symbolic regression (SCSR)~\cite{Kronberger2021,Bladek2019} and (3)
eXtreme gradient boosting (XGBoost)~\cite{chen2016xgboost}. We compare classification
accuracy, supported constraint types, and runtime performance based on data stemming from
a use-case in the automotive industry.

\section{SC-based data validation}
\label{sec:SCBasedDataValidation}
ML algorithms have long been applied for the purpose of data validation. Concept drift
detection~\cite{gama2004learning} applies e.g. ML models and analyzes the prediction error
to detect changes in system behavior. These models are either trained on data
from a manually validated baseline and detect subsequent deviations from this established
baseline, or are trained continuously to detect deviations from previous
states~\cite{Gama2014}. SC-based data validation, however, is able to assess the quality
of unseen data without established baselines by using domain knowledge.

Quality of data is often assessed by analysis of the interaction of inputs values in
regard to the target. The measured target must exhibit certain shape properties that we
associate with \emph{valid} data and \emph{valid} interactions. SCR allows us to train
prediction models on the potentially erroneous dataset, whilst enforcing a set of shape
constraints. Therefore, the trained prediction model exhibits a higher error if the data
contains outliers or erroneous segments that \emph{violate} the provided constraints, as
SCR is restricting the model from fitting to these values.

\begin{figure}
  \includegraphics[width=12cm]{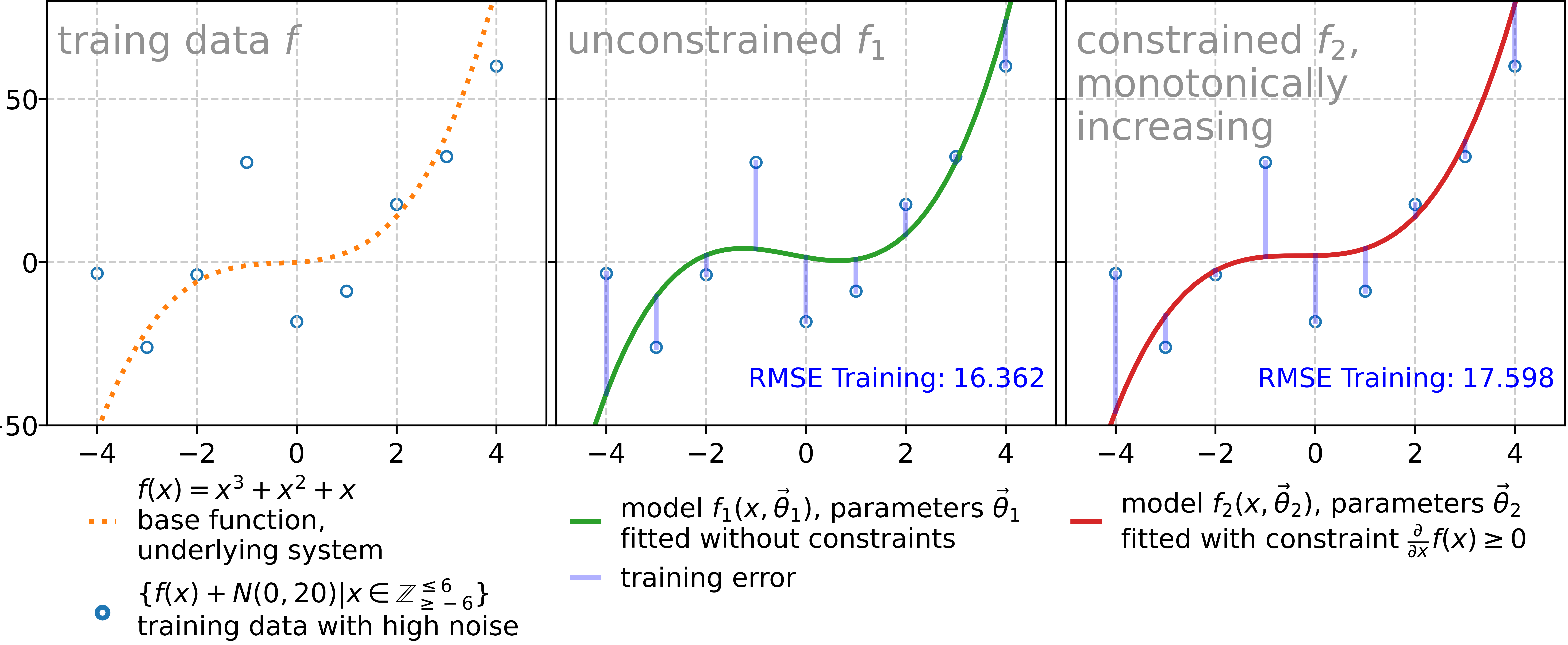}
  \caption{Over-simplified showcase of SC-based data validation. The constrained model $f_2$ exhibits a higher error as it is restricted from fitting to certain data points, but it exhibits the monotonicity of the generating base function.}
  \label{fig:SimplifiedShowcase}
\end{figure}

Figure~\ref{fig:SimplifiedShowcase} shows a simplified example where we sample training
data from a third degree polynomial base function $f$, with added normally distributed
noise. Subsequently, we train the linear factors $\ora{\theta}$ of two third degree
polynomials, with and without constraints. The constrained model $f_2$ exhibits a higher
training error as it includes no decreasing area $x \in [-1,1]$ visible in $f_1$, but exhibits the monotonicity of the generating function as enforced by the constraints.

SC-based data validation requires two prerequisites: (1) precise constraints that describe
\emph{valid} system behavior and (2) a small set of manually validated data. These
manually validated datasets are required to perform a one-time grid search to determine
the best algorithm parameters for each application scenario. Later, for all arriving
unseen datasets, a constrained model is trained on the full data. Similar to
Figure~\ref{fig:SimplifiedShowcase}, datasets are labeled as \emph{invalid} when the model
exhibits a high training error and exceeds a threshold $t$.

\section{Shape-Constrained Regression (SCR)}
Shape-constrained regression (SCR) allows the enforcement of shape-properties of the
regression models. Shape-properties can be expressed as restrictions on the partial
derivatives of the prediction model that are defined for a range of the input space. This
side information is especially useful when training data is limited. The combination of
data with prior knowledge can increase trust in model predictions\cite{cozad2015combined},
which is an equally important property in data validation.
Table~\ref{tab:shape-constraints} lists common examples of shape constraints together with
the mathematical expression and compares the capabilities of the different algorithms.

\subsection{Shape-Constrained Polynomial Regression (SCPR)}
For regular polynomial regression (PR), a parametric (multi-variate) polynomial is fit to
data. This is achieved by fitting the linear coefficients of each term using ordinary
least squares (OLS). For SCPR, we include sum-of-squares constraints (a relaxation of the
shape constraints) to the OLS objective function, which leads to a semidefinite
programming problem (SDP)~\cite{parrilo2000structured}. We use the commercial solver
Mosek\footnote{\url{https://www.mosek.com}} to solve the second-order cone problem (SOCP)
without shape constraints and the SDP with shape constraints. The algorithm parameters of
PR and SCPR are: $d$ the (total) degree of the polynomial, $\lambda$ the strength of
regularization, and $\alpha$ used to balance between 1-norm (lasso regression) and 2-norm
(ridge regression) penalties. SCPR is able to incorporate all constraints of
Table~\ref{tab:shape-constraints}, is deterministic and produces reliable results in
relatively short runtime.

\subsection{Shape-constrained symbolic regression (SCSR)}
SCSR~\cite{Kronberger2021} uses a single objective genetic algorithm (GA) to train a
symbolic regression model. After evaluation, in an additional model selection step, the
constraints are asserted by calculating the prediction intervals on partial derivatives of
the model. Any prediction model that violates a constraint is assigned the error of the
worst performing individual, thereby preserving genetic material. Due to the probabilistic
nature of the GA the achievement of constraints is not guaranteed.

\subsection{XGBoost - eXtreme Gradient Boosting}
XGBoost~\cite{chen2016xgboost} builds an ensemble of decision trees with constant valued
leaf nodes. It is able to consider monotonic constraints, however, these constraints can
only be enforced on the whole input space of one input vector  $\ora{x_i}$. It provides no
support for larger intervals (extrapolation guidance), or multiple (overlapping)
constraint intervals, like SCPR or SCSR. This results in fewer, less specific constraints
available for XGBoost as summarized Table~\ref{tab:shape-constraints}. XGBoost uses the
parameters $\lambda$, $\alpha$ to determine the 1-norm (lasso regression) and 2-norm
(ridge regression) penalties respectively.

\begin{table}[h]
  \center
  \caption{Examples of shape constraints. All constraints marked with $\bullet$ are
  enforced for a domain $[l_i, u_i] \subseteq \ora{x_i}$ of the full $\ora{x_i}$ input
  space. Multiple constraints can be defined over several partitions. Constraints for
  algorithms marked with $\ast$, however, can only be asserted on the full input space of
  $\ora{x_i}$. Constraints marked with $\circ$ are not available.}
  \label{tab:shape-constraints}
  \begin{tabular}{lcccc}
      Property & Mathematical formulation & SCPR & SCSR & XGBoost \\
      Positivity & $f(X) \geq 0$ & $\bullet$ & $\bullet$ & $\circ$  \\
      Negativity & $f(X) \leq 0$ & $\bullet$ & $\bullet$ & $\circ$ \\
      Monotonically increasing & $\frac{\partial}{\partial x_i}f(X) \geq 0$ &
      $\bullet$ & $\bullet$ & $\ast $ \\
      Monotonically decreasing & $\frac{\partial}{\partial x_i}f(X) \leq 0$& $\bullet$
      & $\bullet$ & $\ast$ \\
      Convexity & $\frac{\partial^2}{\partial x_i^2}f(X) \geq 0$& $\bullet$ & $\bullet$ &
      $\circ$  \\
      Concavity & $\frac{\partial^2}{\partial x_i^2}f(X) \leq 0$& $\bullet$ & $\bullet$ &
      $\circ$  \\
  \end{tabular}
\end{table}

\section{Experiment and Setup}
This section provides a short description of the data from our real-world use-case and
discusses the experiment setup used to compare the investigated SCR algorithms based on
this use-case. We follow the general description of SC-based data validation as described
in Section~\ref{sec:SCBasedDataValidation}.

\subsection{Problem Definition - Data from Friction Experiments}
Miba Frictec GmbH\footnote{\url{https://www.miba.com}} develops friction systems such as
breaks or clutches for the automotive industry. The exact friction characteristics of
novel material compositions are unknown during development, and can only be determined by
time- and resource-intense experiments. For this purpose, a friction disc prototype is
installed in room filling test-rigs that rotate the discs at different velocities $v$, and
repeatedly engage the discs at a varying pressure $p$ to simulate the actuation of a
clutch during shifting. Based on these measurements, the friction characteristics of new
discs are determined. The friction coefficient $\mu$ denotes the ratio of friction force
and normal load. It describes the force required to initiate and to maintain relative
motion (denoted static friction $\mu_\textit{stat}$ and dynamic friction
$\mu_\textit{dyn}$)~\cite{IntroductionToTribology2013}. The value of $\mu$ is not constant
for one friction disc, instead, it is dependent on the parameters: $p$, $v$ and
temperature $T$. Experts determine the quality of data by analyzing the interactions of
$p, v, T$ with regard to $\mu_\textit{dyn}$.

In friction experiments we encounter several known issues that render whole datasets or
segments erroneous and that are only detectable when we investigate the interaction of
inputs with regard to the target  $\mu_\textit{dyn}$. Examples for such errors include:
wrong calibration or malfunction of sensors, loosened or destroyed friction pads, or
contaminated test benches from previous failed experiments. We were provided a total of 53
datasets consisting of 18 manually validated and 35 known invalid datasets that were
annotated with a description of the error type.

\begin{align}
  \label{eq:cf_dyn_constraints}
  \nonumber &\forall_{v,p,T} \hskip0.5em v \in [0,1] \wedge p \in [0,1] \wedge T \in [0,1] \implies \\
  \nonumber &\Big( 0 \leq \mu_\textit{dyn} \leq 1 \wedge  \frac{\partial \mu_\textit{dyn}}{\partial v} \in [-0.01,0.01]  \\
  &\wedge \frac{\partial \mu_\textit{dyn}}{\partial p} \leq 0 \wedge \frac{\partial^2 \mu_\textit{dyn}}{\partial p^2} \geq 0
  \wedge \frac{\partial \mu_\textit{dyn}}{\partial T} \leq 0 \wedge \frac{\partial^2 \mu_\textit{dyn}}{\partial T^2} \geq 0 \Big) \\
  &\forall_{p,T} \hskip0.5em p \in [0,1]   \wedge T \in [0,1]  \implies
  \Big( \frac{\partial \mu_\textit{dyn}}{\partial p} \leq 0
  \wedge \frac{\partial \mu_\textit{dyn}}{\partial T}  \leq 0  \Big)
  \label{eq:xgboost_constraints}
\end{align}

\subsection{Experiment Setup}
We performed a hyper-parameter search using a two-fold cross validation over all valid
datasets, repeated for each algorithm. As the constraints define expected behavior, each
algorithm should be able to train models with low test error on valid data, whilst
adhering to the constraints. Equation~\ref{eq:cf_dyn_constraints} lists the constraints
for $\mu_\textit{dyn}$, which were provided by domain experts. Inputs $p, v, T$ are
individually scaled to a range of $[0,1]$ and all constraints are defined for this full
input space. Equation~\ref{eq:xgboost_constraints} lists the reduced constraints that are
compatible with XGBoost's capabilities.

\begin{figure}
  \includegraphics[width=9.5cm]{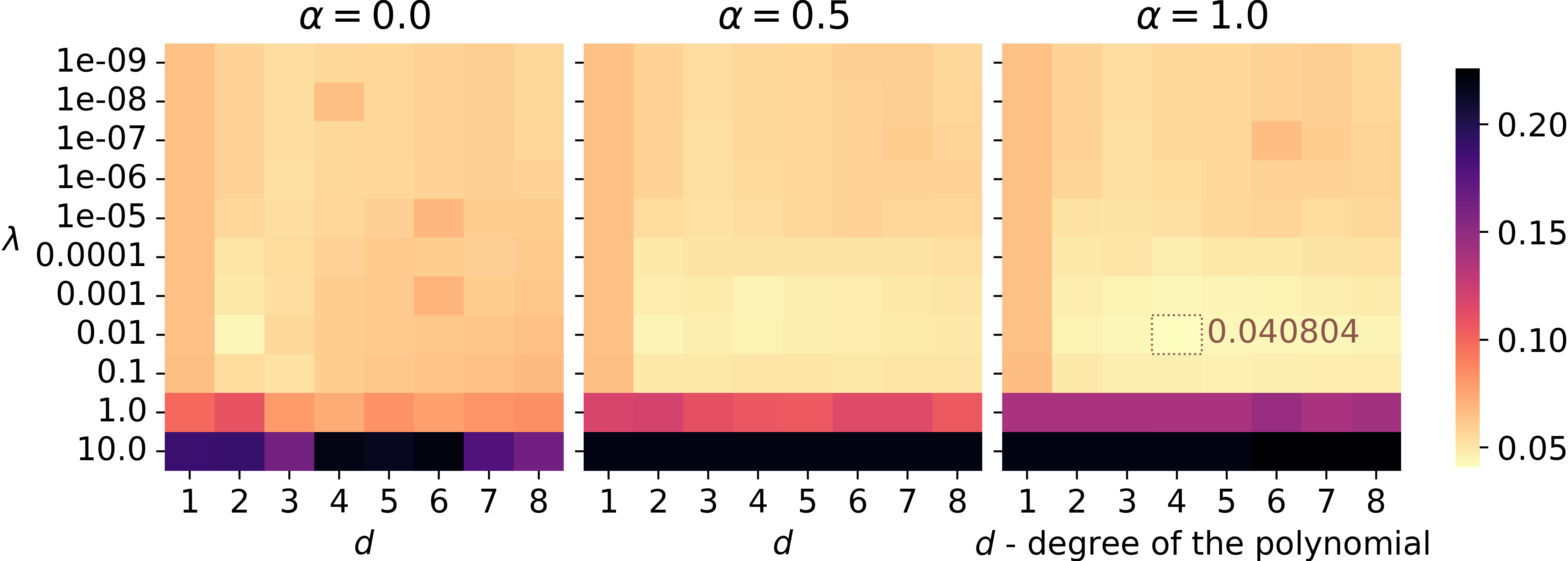}
  \caption{Grid search results for SCPR over all validated datasets. Lower values signify
  better results.}
  \label{fig:SCPRGridsearch}
\end{figure}

Figure~\ref{fig:SCPRGridsearch} visualizes the search space and best configuration for
SCPR in a heatmap showing the sum of test RMSE over all valid datasets. Similar
experiments and analysis were conducted for PR and XGBoost. For SCSR, we compared training
and test error over increasing generation count. To prevent overfitting we select the
generation with the lowest test error as a stopping criterion. In all subsequent training
on unseen data, during the validation phase, the GA is stopped at this generation.

\begin{figure}[h]
  \includegraphics[width=12cm]{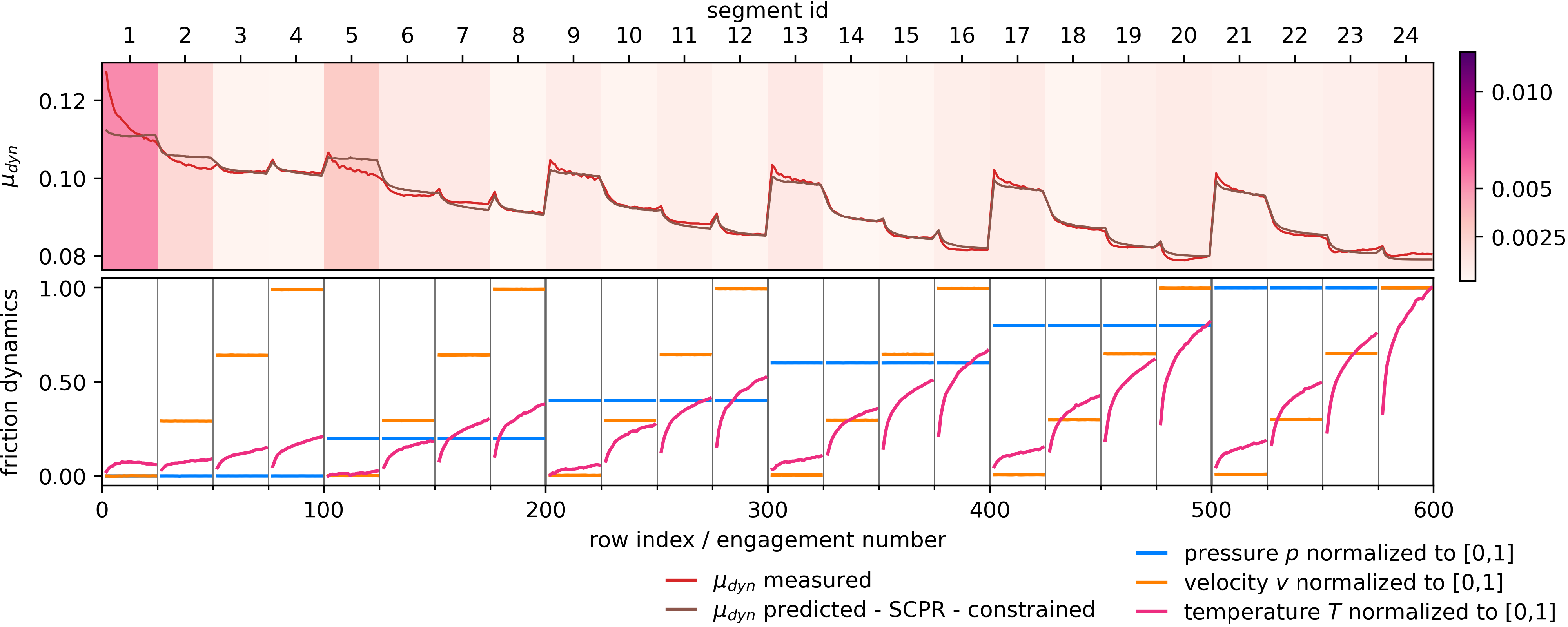}
  \caption{ Validation result for data of one friction experiment. SC-based data
  validation was able to detect the subtle deviations from expected behavior in the
  segments one and five with row IDs 0-25, 100-125. The drops in
  $\mu_\textit{dyn}$-measured are not motivated by the friction dynamics. This dataset was
  correctly labeled \emph{invalid}.}
  \label{fig:ResultExample}
\end{figure}

The resulting best SCR algorithm parameters were applied in the SC-based data validation
phase for all available datasets. In this use-case, we divide the dataset into a new
segment when one of the controlled input parameters $p$ or $v$ changed (cf.
Figure~\ref{fig:ResultExample}). We calculate the RMSE values per segment and mark the
whole experiment as invalid if one segment exceeds the varied threshold $t$.

\section{Results}
\begin{figure}
  \centering
  \hfill
  \subfigure[Receiver operator characteristics (ROC) curve, showing the classification
  capabilities.]{\includegraphics[width=6.39cm]{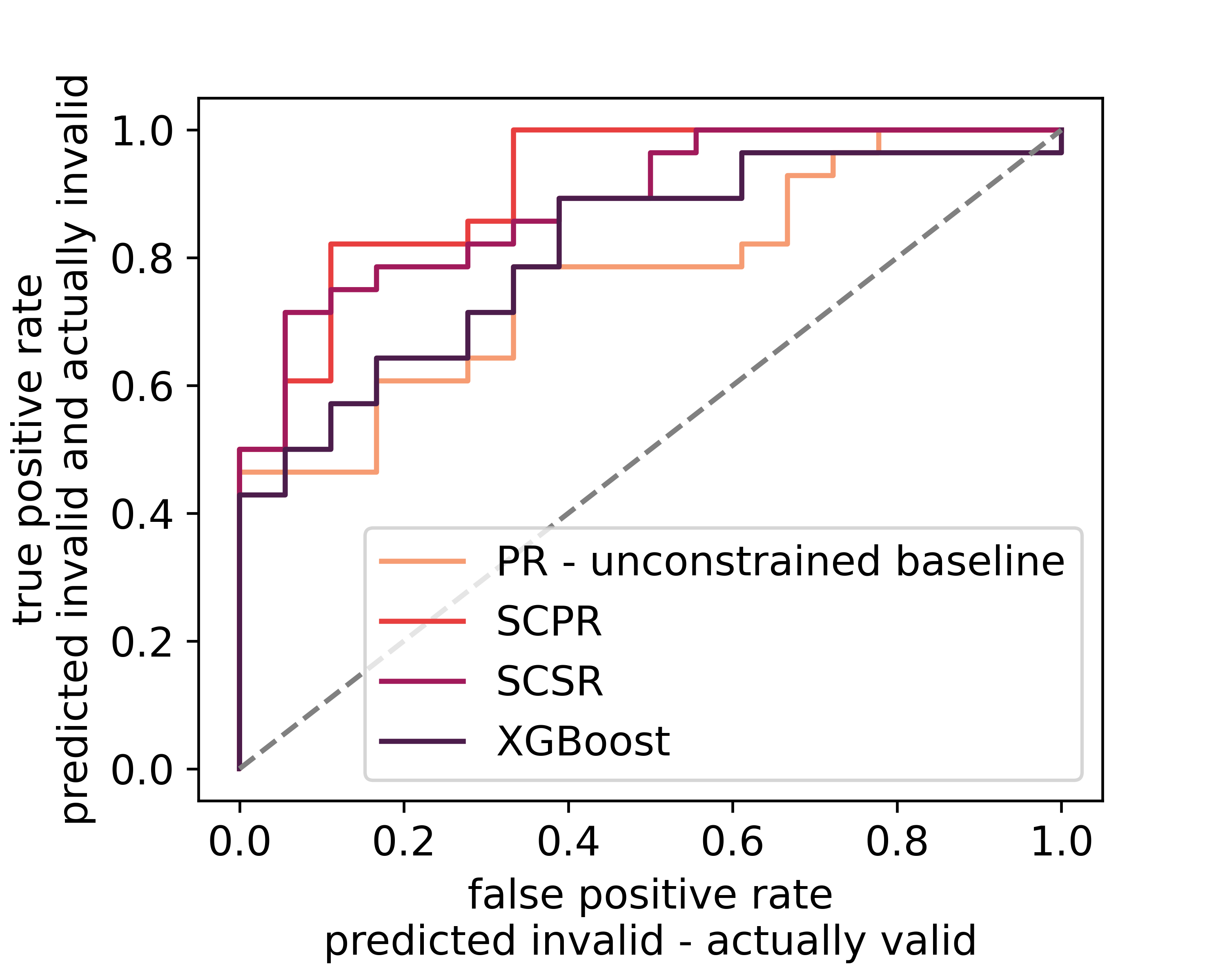}}
  \hfill
  \subfigure[Grid search test error.]{\includegraphics[width=2.80cm]{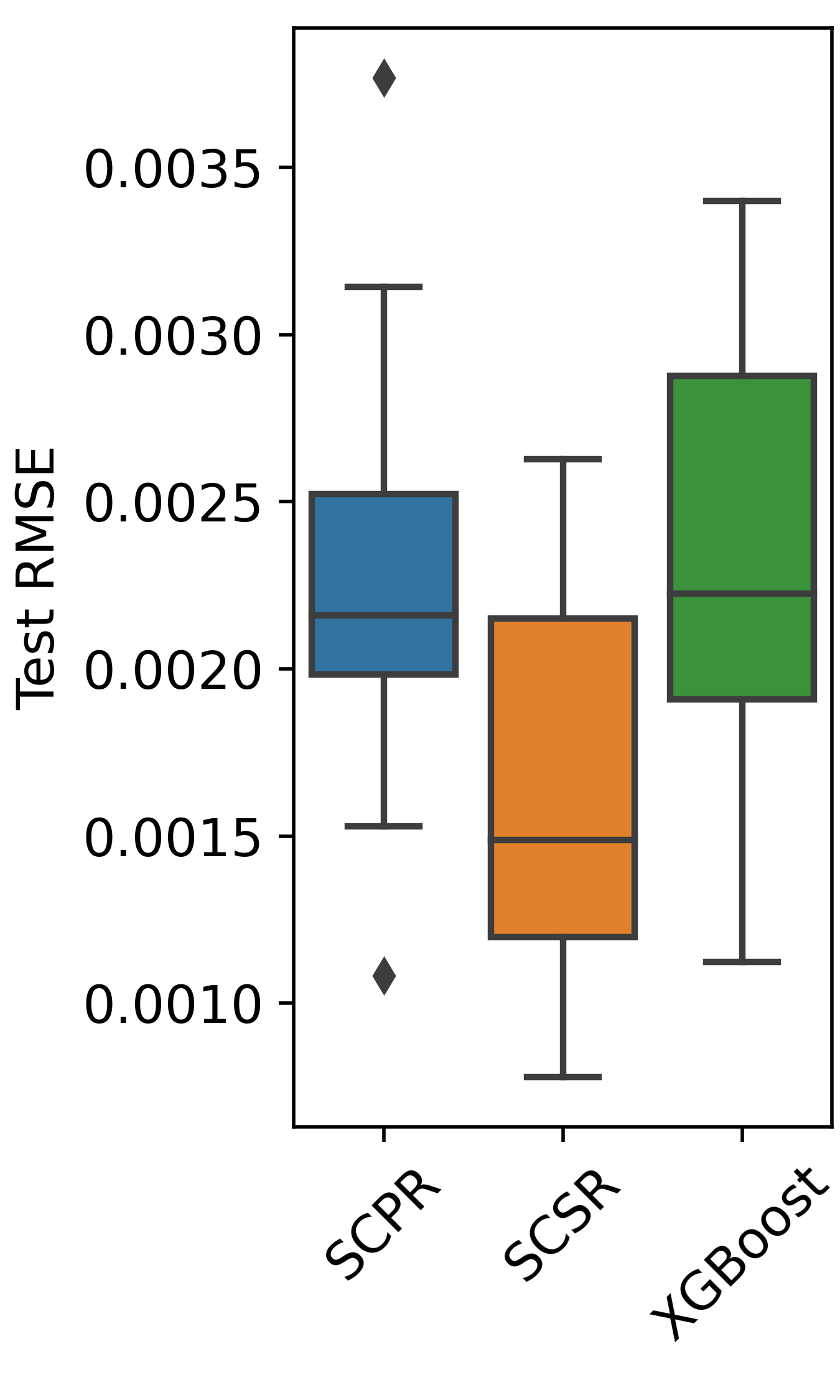}}
  \hfill
  \subfigure[Training duration.]{\includegraphics[width=2.80cm]{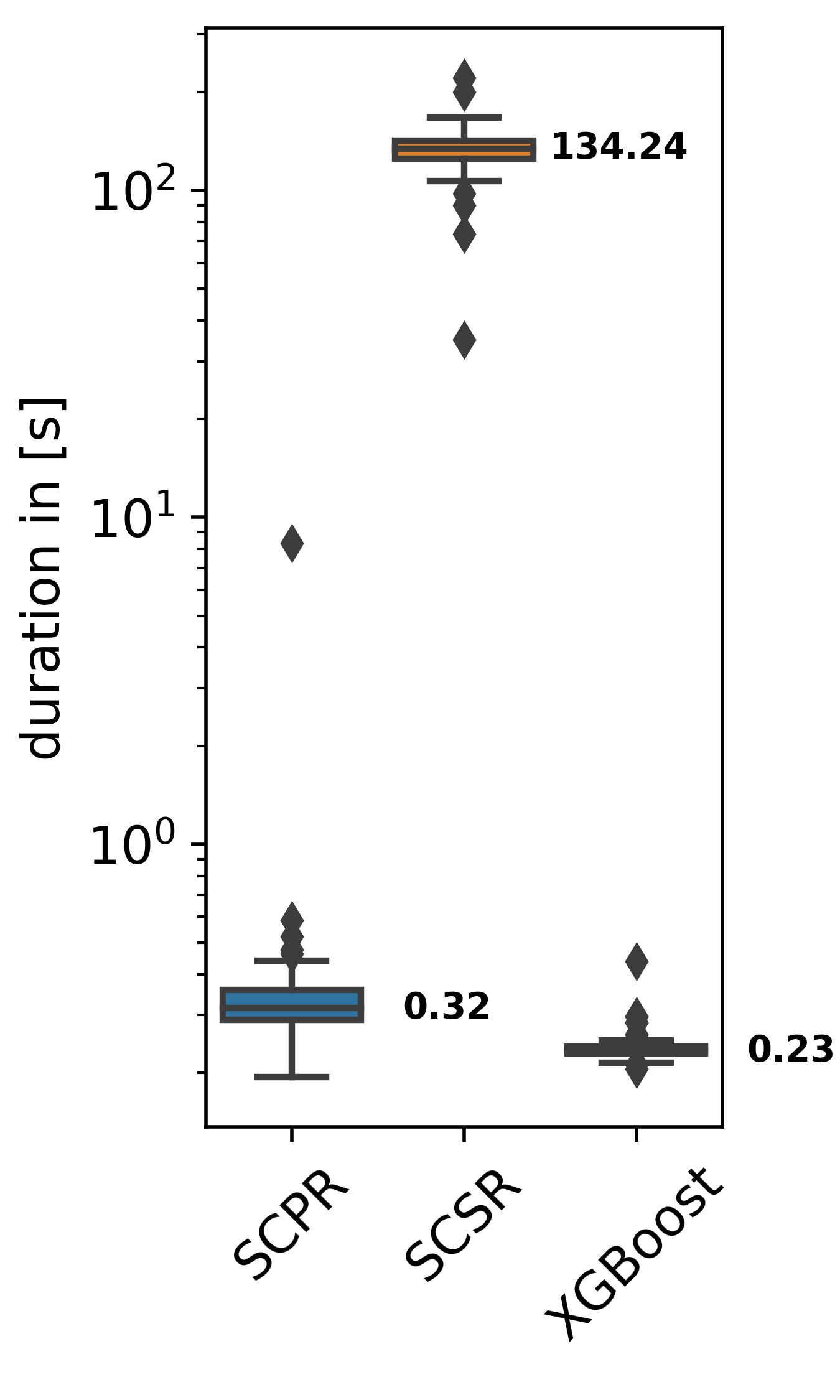}}
  \hfill
  \caption{Performance comparison of the three SCR algorithms and unconstrained PR.}
  \label{fig:algorithmComparison}
\end{figure}

A comparison of the investigated SCR algorithms is visualized in
Figure~\ref{fig:algorithmComparison}. \mbox{XGBoost} supports fewer, less complex shape
constraints and achieves only minimally better classification capabilities than the
unrestricted PR baseline. The comparison with PR shows how many erroneous datasets are
simply detectable due to the statistical properties of ML models. The objective function
of minimized training error leads to models being fit to the behavior represented in the
majority of the data, resulting in the detection of less represented behavior or outliers.
SCPR and SCSR on the other hand exhibit significantly improved classification
capabilities, which can be attributed to the increased restrictions added by the
constraints and domain knowledge about expected \emph{valid} behavior.

We subsequently varied the threshold value $t$ to analyze the change in false-positive-
and true-positive-rate as visualized Figure~\ref{fig:algorithmComparison}. Higher values
of $t$ result in the detection of only severe errors and a lower false positives rate.
Lower values of $t$ cause a more sensitive detection and higher false positive rates.

The sharp vertical incline in the ROC-curve of Figure~\ref{fig:algorithmComparison} is
caused by the numerous \emph{invalid} datasets that exhibit severe errors like e.g.
massive outliers. Such errors cause high training error regardless if constraints are
applied and how restrictive they are. Eventually, for increasingly smaller values of $t$,
even noise present in the data will result in a training error that exceeds $t$.

Figure~\ref{fig:algorithmComparison} also compares the test RMSE values achieved by the
best algorithm parameters on the 18 valid datasets. All three algorithms are similarly
well suited for modeling friction data. Consequently, all conclusions about the data
validation capabilities of individual algorithms are not biased by the training accuracy.
With an average training time of 0.32s per dataset and great classification capabilities,
SCPR is best suited for SC-based data validation. In practical applications, the data
quality assessment is implemented in automated data ingestion pipelines that require low
latencies. SCPR adds only little in terms of computational effort but provides significant
improvement in data quality.

\section{Conclusions}
SC-based data validation is a novel approach that allows the inclusion of prior knowledge
in the quality assessment of previously unseen datasets. It can detect faults in the data
that are only identifiable in the interaction of observables. With its low average
runtime, SC-based data validation using SCPR is suitable for integration into data import
pipelines to improve data quality. Moreover, trust in the validation results is
facilitated by readable constraint definitions that can be provided by domain experts, or
derived from expert knowledge. This trust is further increased through interpretable
models created by the white- or gray-box ML algorithms SCPR and SCSR.

Based on our experiments, we recommend the application of SCPR for SC-based data
validation. SCPR is easy to configure and excels in runtime time performance, as well as
classification accuracy. For cases with larger number of variables or categorical data,
XGBoost might be better equipped.

\subsection*{Acknowledgement}
The financial support by the Christian Doppler Research Association, the
Austrian Federal Ministry for Digital and Economic Affairs and the National
Foundation for Research, Technology and Development is gratefully acknowledged.

\bibliographystyle{splncs}
\bibliography{bachinger}

\end{document}